\documentclass{article}
 
\usepackage[nonatbib,preprint]{neurips_2021}

\usepackage[utf8]{inputenc} 
\usepackage[T1]{fontenc}    
\usepackage{hyperref}       
\usepackage{url}            
\usepackage{booktabs}       
\usepackage{amsfonts}       
\usepackage{nicefrac}       
\usepackage{microtype}      
\usepackage{xcolor}         
\usepackage{algorithm}
\usepackage{bm}
\usepackage{algpseudocode}
\usepackage{amsthm,amsmath,amssymb,amsfonts}
\usepackage{graphicx}

\title{Wasserstein Adversarially Regularized\\ Graph Autoencoder}

\author{%
  Huidong~Liang\\
  Discipline of Business Analytics\\
  The University of Sydney Business School\\
  The University of Sydney\\
  Sydney, NSW 2006, Australia \\
  \texttt{hlia0714@uni.sydney.edu.au} \\
  \And
  Junbin~Gao \\
  Discipline of Business Analytics\\
  The University of Sydney Business School\\
  The University of Sydney\\
  Sydney, NSW 2006, Australia \\
  \texttt{junbin.gao@sydney.edu.au} \\
}

\begin{document}

\maketitle

\begin{abstract}
This paper introduces Wasserstein Adversarially Regularized Graph Autoencoder (WARGA), an implicit generative algorithm that directly regularizes the latent distribution of node embedding to a target distribution via the Wasserstein metric. The proposed method has been validated in tasks of link prediction and node clustering on real-world graphs, in which WARGA generally outperforms state-of-the-art models based on Kullback-Leibler (KL) divergence and typical adversarial framework.
\end{abstract}

\section{Introduction}
Graph, consisting of a set of nodes and links, is an essential type of data that captures the topological structure within observations. Typical tasks on graphs involve link prediction, node clustering, node classification, etc., which have received an increasing amount of attention in social network analysis~\cite{backstrom2011supervised}, recommendation system~\cite{koren2009matrix} and bioinformatics~\cite{agrawal2018large}.

Recently, \textit{node embedding}, an approach that provides low-dimensional vector-space representations for graph nodes, has become a paradigm in graph analysis. The learned embedding preserves useful information from the original features and retains the topological structure of the graph in the meantime, such that the algorithm will be both effective and efficient when performing the aforementioned downstream tasks. Some embedding methods assume that only topological structure is given, such as Spectral Clustering~\cite{tang2011leveraging}, DeepWalk~\cite{perozzi2014deepwalk} and node2vec~\cite{grover2016node2vec}, while others not only consider the graph structure but also exploit the content features, such as Graph Convolutional Networks (GCN)~\cite{kipf2017semi}, GraphSAGE~\cite{hamilton2017inductive} and Graph Attention Networks (GAT)~\cite{velivckovic2017graph}.

Among the embedding models, some take the generative approach, which, instead of learning a fixed vector representation for each node, assume that the latent representation follows a particular probability distribution. For example, Variational Graph Autoencoder (VGAE)~\cite{kipf2016variational} takes a standard Gaussian as the prior distribution, and then uses variational inference to estimate the posterior distribution of the latent embedding, with Kullback-Leibler (KL) divergence being the measurement of ``distance'' between distributions (Nevertheless KL divergence is not a metric for distance by strict definition as it fails to satisfy symmetry and triangular inequality). Adversarially Regularized Graph Autoencoder (ARGA)~\cite{pan2018adversarially}, based on VGAE, further proposes an adversarial framework that involves a discriminator to distinguish the encoded distribution from the prior, and then optimizes both encoder and discriminator simultaneously as a \textit{minimax} problem, leading to a better result in performance.

Meanwhile, Wasserstein distance, also known as Earth-Mover distance~\cite{rubner1998metric},
has gained popularity in machine learning research for its effective measurement of the distance between distributions. Compared to the other commonly used ``distance'' metrics such as KL divergence and Jensen-Shannon (JS) divergence, Wasserstein distance is suitable for estimating distributions with disjoint supports. For instance, Wasserstein Generative Adversarial Networks (WGAN)~\cite{arjovsky2017wasserstein} replaces the discriminator in Generative Adversarial Nets (GAN)~\cite{goodfellow2014generative} by Wasserstein metric to handle the problems of unstable training and mode collapsing. Moreover, Wasserstein Autoencoder (WAE)~\cite{tolstikhin2018wasserstein} outperforms Variational Autoencoder (VAE)~\cite{kingma2014auto} by enforcing a continuous encoded latent distribution to match the target distribution, as opposed to VAE that individually enforces each observation's latent distribution to match the target distribution.

In this work, we propose Wasserstein Adversarially Regularized Graph Autoencoder (WARGA), which directly regularizes the encoded latent distribution to a target distribution via 1-Wasserstein distance.
Compared to the methods that use KL divergence to measure the similarity of graph embeddings' distributions, Wasserstein distance can provide useful information when the two distributions share no common support. Compared to adversarial methods that regularize the embedding by distinguishing encoded distributions from target distributions as a classification task, Wasserstein distance provides a more natural explanation for regularization by using the distance between distributions other than artificially designing a discriminator.
Empirical studies of link prediction and node clustering on three popular citation networks have been conducted to test our model's performance against KL divergence and adversarial framework methods.

\section{Background and related work}
In this section, we review graph embedding approaches that use KL divergence and give an overview on Wasserstein distance.

\subsection{Variational Graph Autoencoder}
Let ${\bf G} = ({\bf V,A})$ denote a graph consisting of a set of $N$ nodes ${\bf V} = \{ {\bf v}_1,...,{\bf v}_N \}$ with their features ${\bf X} = {\{\bf x}_1,...,{\bf x}_N\}$ and an adjacency matrix $\bf A$ with ${\bf A}_{ij} = 1$ if there is a link between node ${\bf v}_i$ and ${\bf v}_j$, and ${\bf A}_{ij} = 0$ otherwise.

VGAE is optimized to perform link prediction task by maximizing the variational lower bound $\mathcal{L}$:
\begin{equation}\label{eq1}
    \mathcal{L} = \mathbb{E}_{q({\bf z|A,X})}[\log p({\bf A|X,z})] - \text{\bfseries KL}[ q({\bf z|A,X}) || p({\bf z}) ],
\end{equation}
in which $\bf z$ is the encoded latent embedding from GCN~\cite{kipf2017semi} with probability distribution $q({\bf z|A,X})$, $p({\bf z})$ is a prior distribution of $\bf z$ (e.g. standard Gaussian), and $p({\bf A|X,z})$ is the likelihood of reconstructing $\bf A$ given $\bf z$ through an inner-product decoder.

We can view such formulation from a regularization perspective: maximizing the variational lower bound $\mathcal{L}$ is equivalent to minimizing the cross entropy loss and the KL divergence between $q$ and $p$:
\begin{equation}\label{eq2}
    \max{\mathcal{L}} \Longleftrightarrow \min \big \{- \mathbb{E}_{q({\bf z|A,X})}[\log p({\bf A|X,z})] + \text{\bfseries KL}[ q({\bf z|A,X}) || p({\bf z}) ] \big \}.
\end{equation}
Therefore, we can treat the KL divergence as a term that penalizes the encoded distribution $q$ for deviating from the specified prior distribution $p$.

\subsection{Adversarially Regularized Graph Autoencoder}
ARGA further introduces an adversarial model $\mathcal{D}$ that discriminates the samples of encoded latent distribution $q_z$ by generator $\mathcal{G}$ from the samples of specified prior $p_z$. Together with the generator $\mathcal{G}$, ARGA is optimized as a \textit{minimax} problem, where the generator wishes to generate embeddings that baffle the discriminator, while the latter tries to discern the ``fake'' embeddings. The adversarial objective is defined as:
\begin{equation}\label{eq3}
    \min_{\mathcal{G}}\max_{\mathcal{D}} \big \{ \mathbb{E}_{{\bf z} \sim p_z} [\log \mathcal{D}({\bf z})] + \mathbb{E}_{{\bf z}\sim q_z} [\log (1 - \mathcal{D}(\mathcal{G}({\bf A, X})))] \big \}.
\end{equation}
Finally, the generator $\mathcal{G}$ will be iteratively updated by both adversarial objective and variational lower bound $\mathcal{L}$ in Eq (\ref{eq1}) to encode the original feature into a regularized latent embedding.

\subsection{Wasserstein distance and its dual form}
\label{section2.3}
The 1-Wasserstein distance~\cite{villani2021topics} between two distributions $\mathbb{P}_r$ and $\mathbb{P}_g$ is defined as:
\begin{equation}
    W_1(\mathbb{P}_r,\mathbb{P}_g) = \inf_{\gamma \in \mathcal{P} (r \sim \mathbb{P}_r, z \sim \mathbb{P}_g)} \mathbb{E} [\| r - z\|_2],
\end{equation}
where $\mathcal{P} (r \sim \mathbb{P}_r, z \sim \mathbb{P}_g)$ is the set of all joint distributions $\gamma (r,g)$ with marginals $\mathbb{P}_r$ and $\mathbb{P}_g$ respectively. As the above form is intractable, the expression can be reformulated by Kantorovich-Rubinstein duality as:
\begin{equation}\label{eq5}
    W_1(\mathbb{P}_r,\mathbb{P}_g) = \sup_{\|f\|_L \leq 1} \big\{ \mathbb{E}_{r \sim \mathbb{P}_r}[f(r)] - \mathbb{E}_{z \sim \mathbb{P}_g}[f(z)] \big\},
\end{equation}
where $f$ is any continuous function that satisfies 1-Lipschitz continuity. In Wasserstein GAN~\cite{arjovsky2017wasserstein} the authors show that such problem can be solved via:
\begin{equation}
    \max_{\phi \in \Phi} \big\{ \mathbb{E}_{r \sim \mathbb{P}_r}  [f_{\phi}(r)] - \mathbb{E}_{z \sim \mathbb{P}_g}[f_{\phi}(z)] \big\},
\end{equation}
with $f$ parameterised by $\phi$ in a compact space $\Phi$ for $f_{\phi}$ to satisfy 1-Lipschitz constraint. In practice, the parameters $\phi$ can be clipped into a fixed range (e.g. $\Phi = [-0.01,0.01]$) after each iteration during optimization.

\section{Proposed method}
\begin{figure}[H]
  \centering
  \includegraphics[width=12cm]{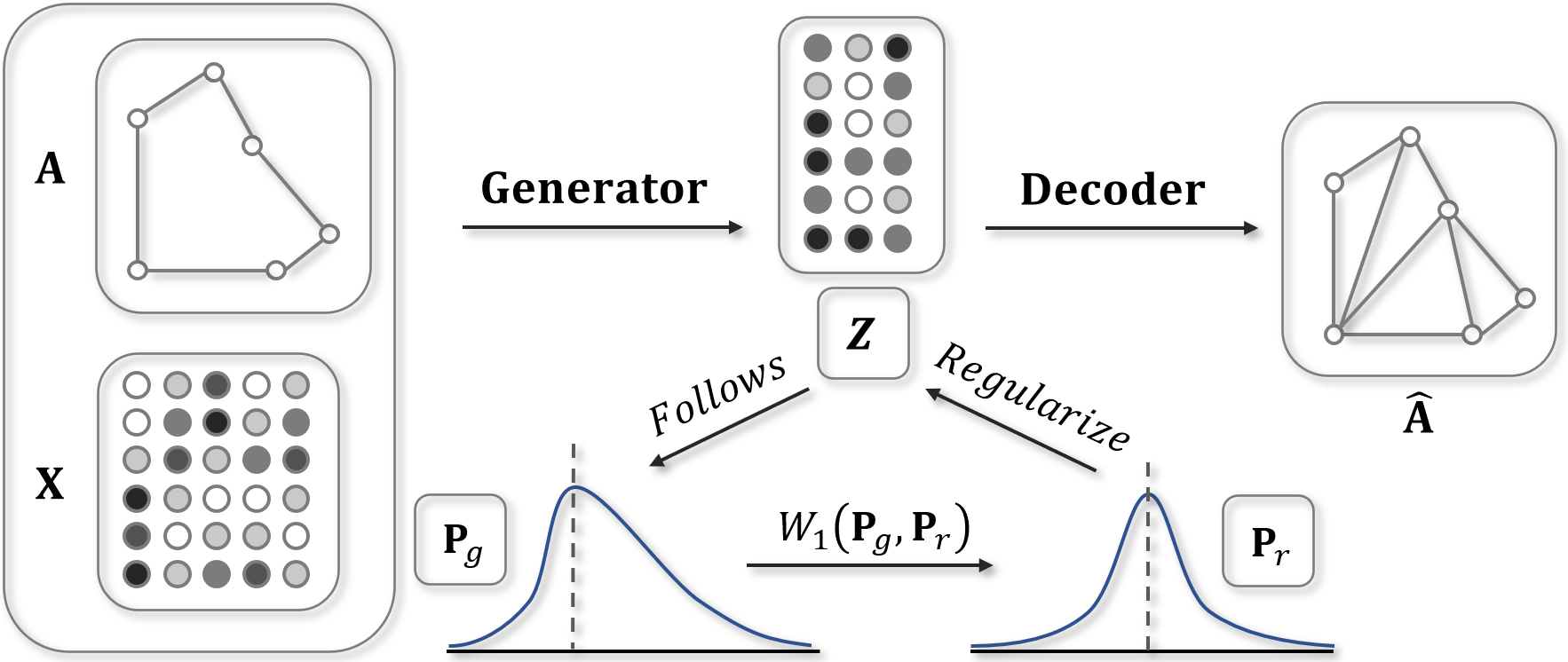}

  \caption{Overall structure of WARGA}\label{fig1}
\end{figure}

The general structure of the proposed method is illustrated in Figure \ref{fig1}. We first use a generator $\mathcal{G}$ to encode the original graph nodes into vector representations, then force the encoded embedding's distribution to match a target distribution by minimizing their 1-Wasserstein distance as regularization. Finally, we reconstruct the adjacency matrix and iteratively update the generator and the regularizer until the algorithm converges.

\subsection{Graph Autoencoder}
\label{section3.1}
We use a 2-layer GCN as generator $\mathcal{G}_w ({\bf A,X})$ to encode the original node features ${\bf X} \in \mathbb{R}^{N \times c}$ with the topological structure $\mathbf A$ into a low dimensional representation ${\bf Z} \in \mathbb{R}^{N \times e}$:
\begin{equation}
    {\mathcal{G}_w ({\bf A}, {\bf X})} = \text{ReLU}( \Bar{\bf A} \, \text{ReLU}(\Bar{\bf A} {\bf X} {\bf W}_1) {\bf W}_2 ), \label{eq7}
\end{equation}
where $\Bar{\bf A} := \Tilde{\bf D}^{-\frac{1}{2}} \Tilde{\bf A} \Tilde{\bf D}^{-\frac{1}{2}}$ is the new "weighted" adjacency matrix for graph $\bf G$ after convolution, with $ \Tilde{{\bf D}}_{{ii}} := \sum_j \Tilde{\bf A}_{_{ij}}$ to be the degree matrix of $\Tilde{\bf A} := \bf A + I$. $\text{ReLU}(t) = \max(0,\,t)$ is used as the activation function in the neural networks with weights ${\bf W}_1 \in \mathbb{R}^{c \times d}$ and ${\bf W}_2 \in \mathbb{R}^{d \times e}$, and the output matrix $\bf Z$ contains latent embeddings ${\bf z}_i$ for each node ${\bf v}_i \in {\bf V}$ as row-vectors.

We assume that the latent representation follows a standard Gaussian distribution $\mathcal{N} (0,{\bf I})$ denoted by $\mathbb{P}_r$, and denote the distribution of the embedding generated by $\mathcal{G}_w$ as $\mathbb{P}_g ({\bf z|A, X})$, where the parameter $w$ is to be learned by reconstructing the adjacency matrix using an inner-product decoder:
\begin{equation}
    p({\bf A|X,Z}) = \prod_{i=1}^n \prod_{j=1}^n p({\bf A}_{ij}|{\bf z}_i,{\bf z}_j), \hspace{0.25 cm} \text{with} \hspace{0.25 cm} p({\bf A}_{ij} = 1|{\bf z}_i, {\bf z}_j) = \sigma({\bf z}_i^\top {\bf z}_j). \label{eq8}
\end{equation}
Here we use $sigmoid$ function $\sigma (t) = 1/(1 + \exp(- t))$ to restrain the output into range (0,1). As such, the objective can be defined as minimizing the cross entropy loss over the parameters $w$ in $\mathcal{G}$ :
\begin{equation}
    \min_{w} - \mathbb{E}_{\hspace{0.07cm} \mathbb{P}_g}[\log p({\bf A|X,Z})].
\end{equation}

\subsection{Wasserstein regularizer}
To force the encoded distribution $\mathbb{P}_g({\bf z|A,X})$ into the target distribution $\mathbb{P}_r = \mathcal{N} (0,{\bf I})$, we introduce a Wasserstein regularizer that helps to minimizes the 1-Wasserstein distance between $\mathbb{P}_r$ and $\mathbb{P}_g$.

From Section~\ref{section2.3} we can formulate the distance as:
\begin{equation}
    W_1(\mathbb{P}_r,\mathbb{P}_g) = \max_{\phi \in \Phi} \big\{ \mathbb{E}_{{\bf r} \sim \mathbb{P}_r} [f_{\phi}({\bf r})] - \mathbb{E}_{{\bf z} \sim \mathbb{P}_g}[f_{\phi}({\bf z})] \big\},
\end{equation}
where $f$ is parameterised by a fully connected neural network with parameters $\phi \in \Phi$ in a compact space. Here we demonstrate $f_\phi$ by a Multilayer Perceptron (MLP) with 2 hidden layers:
\begin{equation}\label{eq11}
    f_{\phi \in \Phi}({\bf z}) = {\bf W}_5\sigma({\bf W}_4\sigma({\bf W}_3{\bf z} + {\bf b}_1) + {\bf b}_2) + {\bf b}_3,
\end{equation}
in which ${\bf W}_3 \in \mathbb{R}^{k \times e}$ and ${\bf W}_4 \in \mathbb{R}^{l \times k}$ are weights in the hidden layers with biases ${\bf b}_1$ and ${\bf b}_2$, while ${\bf W}_5 \in \mathbb{R}^{1 \times l}$ and ${\bf b}_3$ are the parameters in the output layer.

The generator $\mathcal{G}_w$, on the contrary, wants to minimize such distance, which leads to an adversarial-like framework with a \textit{minimax} objective:
\begin{equation}
    \min_{w}\max_{\phi \in \Phi} \big \{ \mathbb{E}_{{\bf r} \sim \mathbb{P}_r} [f_{\phi}({\bf r})] - \mathbb{E}_{{\bf z} \sim \mathbb{P}_g}[f_{\phi}({\bf z})] \big \}.
\end{equation}
Combining with the objective of reconstruction for generator $\mathcal{G}_w$ in Section \ref{section3.1}, we arrive at the final loss function for training:
\begin{equation}\label{eq13}
        \min_{w}\max_{\phi \in \Phi} \big\{ - \mathbb{E}_{ \hspace{0.07cm} \mathbb{P}_g}[\log p({\bf A|X,Z})] + \mathbb{E}_{{\bf r} \sim \mathbb{P}_r} [f_{\phi}({\bf r})] - \mathbb{E}_{{\bf z} \sim \mathbb{P}_g}[f_{\phi}({\bf z})] \big \}.
\end{equation}

\subsection{Learning algorithm}\label{section3.3}
The algorithm for learning the generator and regularizer is demonstrated in Algorithm~\ref{algo1}.

\begin{algorithm}
	\caption{Wasserstein Adversarially Regularized Graph Autoencoder} \label{algo1}
	\textbf{Require}: Graph ${\bf G} = ({\bf V,A})$; feature matrix ${\bf X} \in \mathbb{R}^{N \times c}$; Number of epochs $T$;
	
     \hspace{1.34cm} Number of iterations for Wasserstein regularizer $K$
	
	\begin{algorithmic}[1]
    \For {epoch $= 1, 2, ..., T$}
        \State Encode ${\bf A}$ and ${\bf X}$ into low-dimensional representation ${\bf Z} \in \mathbb{R}^{N \times e}$ by $\mathcal{G}_w$
		\For {iteration $= 1, 2, ..., K$}
		    \State Sample $\{{\bf r}^{i}\}_{i=1}^m$ $\sim \mathbb{P}_r$ a batch of priors
		    \State Sample $\{{\bf z}^{i}\}_{i=1}^m$ $\sim \mathbb{P}_g$ a batch from the encoded embedding
		    \State Update $f_\phi$ by computing:
		    $\nabla_\phi \frac{1}{m}  \sum_{i = 1}^m \big( f_\phi({\bf r}^i) - f_\phi({\bf z}^i) \big)$
		    \State Clip $\phi$ back to $\Phi$
		\EndFor
		
		\State Update $\mathcal{G}_w$ by computing $\nabla_w$ in Eq (\ref{eq13})
	\EndFor\\
	\Return $\bf Z$, $\mathcal{G}_w$ and $f_\phi$
	\end{algorithmic}
\end{algorithm}
The generator $\mathcal{G}_w$ has complexity $\mathcal{O}(Ncde)$ in Eq (\ref{eq7}) as $\Bar{\bf A} {\bf X}$ can be efficiently computed by sparse-dense matrix multiplication~\cite{kipf2017semi}, and the regularizer $f_\phi$ parameterised by a MLP with 2 hidden layers has complexity of $\mathcal{O}(mek + mkl)$ in Eq (\ref{eq11}) if the number of samples $m$ for prior and encoded embedding is much greater than the number of neurons in each hidden layer ($k$ and $l$).

\section{Experiments}
\label{others}
\subsection{Experimental setup}\label{section4.1}
We validate our proposed methods by link prediction and node clustering on three popular citation networks: Cora, Citeseer and PubMed, with their statistics summarized in Table~\ref{tab1}.

\begin{table}[ht]
\centering
\caption{Dataset statistics}\label{tab1}
\setlength{\tabcolsep}{6 pt}
\begin{tabular}{l c c c c}
\hline
{\bfseries Dataset} & \# Nodes & \# Links & \# Features & \# Classes \\
\hline
Cora     & 2,708 & 5,429 & 1,433  & 7\\
Citeseer & 3,327 & 4,732 & 3,703  & 6\\
PubMed   &  19,717 & 44,338 & 500 & 3\\
\hline
\end{tabular}
\end{table}

For both tasks, we compare our algorithms against GAE, VGAE, ARGA and ARVGA. In link prediction, we report AUC score (the area under the curve) and AP score (average precision), while for node clustering, we choose Acc (accuracy), NMI (normalized mutual information) and ARI (adjusted random index) as metrics to compare our results to other baselines. Similar to the experimental design in VGAE~\cite{kipf2016variational}, we use 5\% links for validation, 10\% links for testing, and the rest for training. All experiments are repeated 10 times by different random seeds, with results reported by mean (in percentage) and standard deviation (in decimal). The experiment is conducted on an NVIDIA GEFORCE RTX 3090 GPU, and the codes for replicating the experiment results can be retrieved from \url{https://anonymous.4open.science/r/WARGA-44F6/}.

\subsection{Link prediction}
\subsubsection{Hyper-parameter settings}\label{section4.2.1}
For a fair comparison purpose, we build our encoder identical to other baselines with 32 neurons in the first hidden layer and 16 neurons in the second embedding layer. The Wasserstein regularizer is constructed similar to the discriminator in ARGA with 2 hidden layers (16-neuron and 64-neuron). For Cora and Citeseer, we train our proposed model for 200 epochs via Adam optimizer~\cite{kingma2014adam} and choose a 0.001 learning rate for both encoder and Wasserstein regularizer, with parameters in the latter clamped into [-0.01, 0.01]. While for PubMed dataset, as it is relatively large (around 20$k$ nodes with 44$k$ links) compared to the other graphs, we iterate 1500 epochs for sufficient optimization with a learning rate of 0.005.

\subsubsection{Link prediction results}
We retain the experimental results from ARGA~\cite{pan2018adversarially} for the four baselines together with our results, which are summarized in Table~\ref{tab2}. The results suggest that by incorporating a Wasserstein regularizer, WARGA outperforms all four baselines on Cora and Citeseer with an increase in AUC score and AP score by 0.5\% on average compared to the leading baselines of similar model sizes. While on PubMed dataset, although ARGA achieves the best performance with around 97\% in both AUC score and AP score, our WARGA still generates very competitive results compared to ARGA that are only 0.3\% and 0.1\% lower under AUC and AP respectively.
\begin{table}[h]
\centering

\caption{Link prediction results}\label{tab2}
\setlength{\tabcolsep}{2 pt}
\begin{tabular}{l c c c c c c}
\hline
{\bfseries Method} & \multicolumn{2}{c}{Cora} & \multicolumn{2}{c}{Citeseer} & \multicolumn{2}{c}{PubMed}\\
& AUC & AP & AUC & AP& AUC & AP\\
\hline
{GAE}  & 91.0 $\pm$ 0.02  & 92.6 $\pm$ 0.01 & 89.5 $\pm$ 0.04 & 89.9 $\pm$ 0.05 & 96.4 $\pm$ 0.00 & 96.5 $\pm$ 0.00  \\
{VGAE}  & 91.4 $\pm$ 0.01  & 92.6 $\pm$ 0.01 & 90.8 $\pm$ 0.02 & 92.0 $\pm$ 0.02 & 94.4 $\pm$ 0.02 & 94.7 $\pm$ 0.02  \\
\hline
{ARGA}  & 92.4 $\pm$ 0.003  & 93.2 $\pm$ 0.003 & 91.9 $\pm$ 0.003 & 93.0 $\pm$ 0.003 & {\bf 96.8} $\pm$ {\bf 0.001} & {\bf 97.1} $\pm$ {\bf 0.001}  \\
{ARVGA}  & 92.4 $\pm$ 0.004  & 92.6 $\pm$ 0.004 & 92.4 $\pm$ 0.003 & 93.0 $\pm$ 0.003 & 96.5 $\pm$ 0.001 & 96.8 $\pm$ 0.001  \\
\hline
{WARGA} & {\bfseries 92.9} $\pm$ {\bf 0.003 } &	{\bfseries 93.8} $\pm$ {\bf 0.002 } &	{\bfseries 92.9} $\pm$ {\bf 0.004 } &    {\bfseries 93.6} $\pm$ {\bf 0.004 } &	{ 96.5 $\pm$ 0.001}    &	{ 97.0 $\pm$ 0.001}    \\
\hline
\end{tabular}
\end{table}

\subsubsection{Hyper-parameter analysis}
\begin{figure}[ht]
  \centering
  \includegraphics[width=\textwidth]{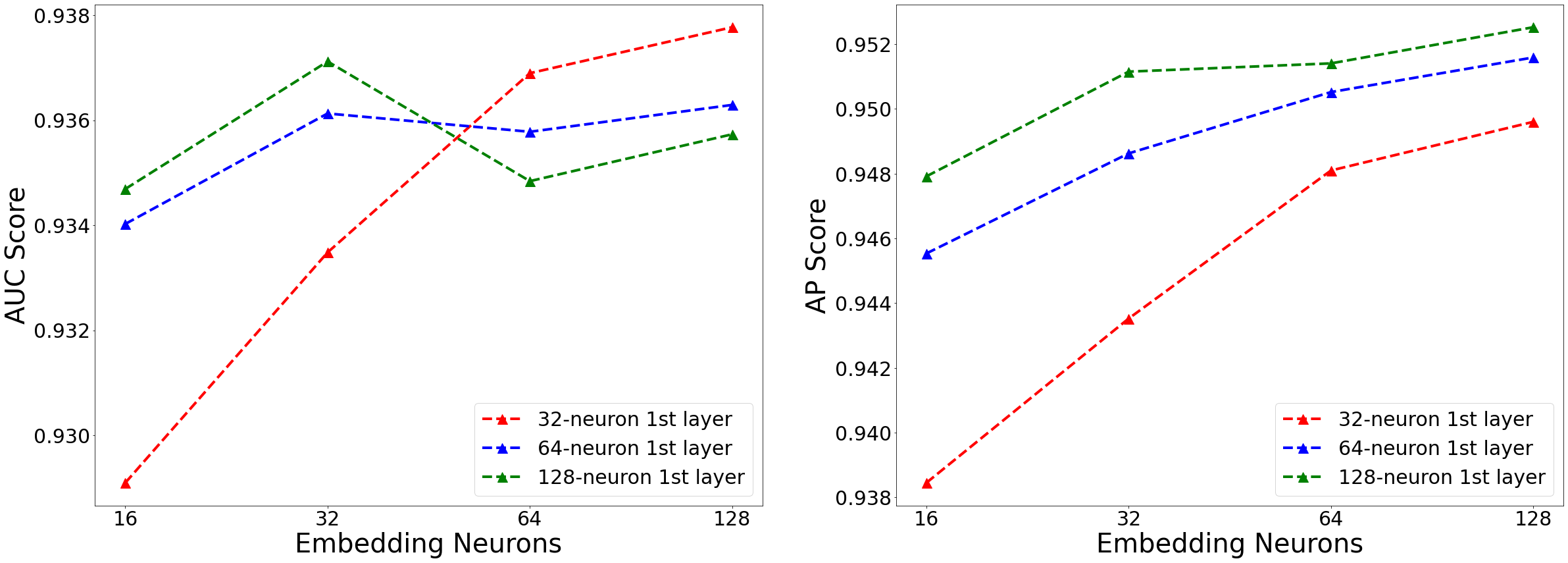}

  \caption{Hyper-parameter analysis on Cora}\label{fig2}
\end{figure}

We further conduct a hyper-parameter analysis to explore the changes in WARGA's performance when given different encoding layers, and demonstrate our findings with Cora dataset. The investigated combinations consist of the first encoding layers chosen from [ 32, 64, 128 ] neurons and the second embedding layers chosen from [ 16, 32, 64, 128 ] neurons. The results are the means of 10 runs with different random seeds, as illustrated in Figure \ref{fig2}.

The results reveal that adding neurons to the second embedding layer when given a first encoding layer with 32 neurons will conduce to conspicuously better performance, but such benefit is diminishing as the number of neurons in the first encoding layer increases. On the other hand, when given a 16-neuron embedding layer, the differences in performance from various encoding layers are significant. However, these gaps also tend to shrink as we increase the embedding neurons from 16 to 128.

\subsection{Node clustering}
In this section, we perform node clustering using K-means algorithm based on the embedding learned from link prediction task above to cluster similar nodes into the same groups. We retain the experiment results for the four baselines from \cite{mrabah2021rethinking} with ours, as shown in Table~\ref{tab3} and Table~\ref{tab4}.

\begin{table}[H]
\centering

\caption{Node clustering results on Cora and Citeseer}\label{tab3}

\setlength{\tabcolsep}{2 pt}
\begin{tabular}{l c c c }
\hline
{\bfseries Cora} & Acc & NMI & ARI\\
\hline
{GAE}   & 55.6 $\pm$ 0.05 & 41.2 $\pm$ 0.03 & 33.2 $\pm$ 0.05 \\
{VGAE}  & 58.6 $\pm$ 0.05 & 40.1 $\pm$ 0.03 & 34.2 $\pm$ 0.03 \\ \hline
{ARGA}  & 59.3 $\pm$ 0.04 & 42.2 $\pm$ 0.03 & 31.6 $\pm$ 0.05 \\
{ARVGA} & 63.4 $\pm$ 0.01 & 45.3 $\pm$ 0.00 & 39.2 $\pm$ 0.02 \\
\hline
{WARGA} & {\bf 66.0 }$\pm$ {\bf 0.03} & {\bf 49.0} $\pm$ {\bf 0.02} & {\bf 43.8} $\pm$ {\bf 0.02} \\
\hline
\end{tabular}
\hfill
\begin{tabular}{l c c c }
\hline
{\bfseries Citeseer} & Acc & NMI & ARI\\
\hline
{GAE}   & 42.5 $\pm$ 0.05 & 19.9 $\pm$ 0.03 & 13.7 $\pm$ 0.06\\
{VGAE}  & 50.3 $\pm$ 0.02 & 23.6 $\pm$ 0.02 & 22.1 $\pm$ 0.02 \\
\hline
{ARGA}  & 36.6 $\pm$ 0.08 & 28.4 $\pm$ 0.04 & 16.1 $\pm$ 0.08  \\
{ARVGA} & 51.5 $\pm$ 0.03 & 26.3 $\pm$ 0.01 & 22.7 $\pm$ 0.02  \\
\hline
{WARGA} & {\bf 56.2} $\pm$ {\bf 0.03} & {\bf 30.1} $\pm$ {\bf 0.02} & {\bf 28.5} $\pm$ {\bf 0.02} \\
\hline
\end{tabular}
\end{table}

Similar to link prediction results, our proposed method outperforms all baselines on Cora and Citeseer datasets in every metric by a decent margin of around 3\% to 5\%. For PubMed dataset, however, VGAE shows the best results under Acc and ARI metrics of 68.9 and 30.6 respectively, while our WARGA achieves the best NMI score of 29.4, with slightly lower Acc and ARI scores of 67.4 and 28.5 compared to the best baseline.
\begin{table}[H]
\centering

\caption{Node clustering results on PubMed}\label{tab4}

\setlength{\tabcolsep}{2 pt}
\begin{tabular}{l c c c }
\hline
{\bfseries PubMed} & Acc & NMI & ARI\\
\hline
{GAE}   & 63.7 $\pm$ 0.01 & 23.3 $\pm$ 0.01 & 22.7 $\pm$ 0.02\\
{VGAE}  & {\bf 68.9} $\pm$ {\bf 0.01} & 28.3 $\pm$ 0.01 & {\bf 30.6} $\pm$ {\bf 0.01} \\
\hline
{ARGA}  & 68.0 $\pm$ 0.00 & {\bf 29.4} $\pm$ {\bf 0.02} & 29.3 $\pm$ 0.00 \\
{ARVGA} & 63.4 $\pm$ 0.00 & 23.1 $\pm$ 0.00 & 22.4 $\pm$ 0.00 \\
\hline
{WARGA} & 67.4 $\pm$ 0.01 & {\bf 29.4} $\pm$ {\bf 0.02} & 28.5 $\pm$ 0.02 \\
\hline
\end{tabular}
\end{table}

\section{Limitation}\label{section5}
Although WARGA shows competitive results in the above experiments, there is still some minor limitation in the model design. The Kantorovich-Rubinstein dual for 1-Wasserstein distance in Eq~(\ref{eq5}) requires supreme over all functions that satisfy 1-Lipschitz continuity. However, the practical approach we adopted from GAN ~\cite{arjovsky2017wasserstein} of clipping parameters $\phi$ into a compact space $\Phi$ only searches within a subspace of the defined space, hence serving as an approximation rather than strictly ensuring the constraint.

\section{Conclusion}
In this work, we introduced Wasserstein Adversarially Regularized Graph Autoencoder, which enforces graph node's latent representation to follow a target distribution by measuring Wasserstein distance between distributions. Empirical results show that the proposed method generally outperforms KL divergence ``regularized'' and typical adversarially regularized methods in link prediction and node clustering. Potential limitation is discussed in the end and future work may explore more on different architectures and generators for graph embedding with Wasserstein regularizer.
\bibliographystyle{plain}

\end{document}